\documentclass[conference]{IEEEtran}
\IEEEoverridecommandlockouts
\usepackage{cite}
\usepackage{amsmath,amssymb,amsfonts}
\usepackage{algorithmic}
\usepackage{graphicx}
\usepackage{textcomp}
\usepackage{xcolor}
\usepackage{tikz}
\usepackage[hidelinks]{hyperref}
\usepackage{float}
\usetikzlibrary{arrows.meta}
\def\BibTeX{{\rm B\kern-.05em{\sc i\kern-.025em b}\kern-.08em
    T\kern-.1667em\lower.7ex\hbox{E}\kern-.125emX}}
\begin{document}

\title{Toward On-Device Markerless 6DoF Tracking of a Physical Manipulative on a Mixed-Reality Headset: A Preliminary Study\\

\thanks{This material is based upon work supported by the National
Science Foundation under Award No.~2119549.}
}

\author{\IEEEauthorblockN{1\textsuperscript{st} Guangxi Feng}
\IEEEauthorblockA{\textit{Department of Architecture} \\
\textit{Texas A{\&}M University}\\
College Station, Texas \\
feng.g@tamu.edu}
\and
\IEEEauthorblockN{2\textsuperscript{nd} Philip Yasskin}
\IEEEauthorblockA{\textit{Department of Mathematics} \\
\textit{Texas A{\&}M University}\\
College Station, Texas \\
yasskin@tamu.edu}
\and
\IEEEauthorblockN{3\textsuperscript{rd} Wei Yan}
\IEEEauthorblockA{\textit{Department of Architecture} \\
\textit{Texas A{\&}M University}\\
College Station, Texas \\
wyan@tamu.edu }

}

\maketitle

\begin{abstract}
AR-Classroom uses physical models and virtual overlays to help learners
explore spatial transformations. To align these overlays on a standalone
headset, we developed a markerless six-degree-of-freedom (6DoF) tracker
for a LEGO airplane on the Meta Quest~3. A compact YOLO model trained
on synthetic Unity renders predicts 20 keypoints from dynamic RGB crops
resized to 128 by 128 pixels. A perspective-n-point solver estimates
the airplane's pose from these predictions and the corresponding 3D
landmarks. In a paired evaluation on 1,000 synthetic validation images,
cropping guided by ground-truth boxes increased the detection rate on
annotated images from 32.5 to 46.2 percent. Mean keypoint F1 increased
from 42.9 to 44.6 percent, with lower keypoint precision in the cropped
view. In preliminary headset tests, 2~Hz was the highest tested tracking
target at which the application maintained approximately 72 frames per
second. A headset demonstration illustrates virtual overlay alignment
with the physical airplane before and after a manual rotation. Quantitative 6DoF pose accuracy on real headset footage will be measured in future work
\end{abstract}


\section{Introduction}
\subsection{Application and Motivation}

Manipulating a physical model gives students a way to connect spatial transformations with their mathematical representations. Earlier desktop versions of AR-Classroom used a webcam to display an augmented view of the model on a computer screen. A student could rotate the model about a specified axis and observe the corresponding changes in its coordinate frame, rotation angle, and transformation matrix~\cite{shaghaghian2024spatial}.

The project has since extended this activity to head-mounted mixed reality~\cite{arclassroom}. In a Meta Quest~3 prototype, students can view a physical airplane LEGO model and its virtual representation together as the model moves. In this setting, the overlay must stay aligned as both the model and the student's viewpoint change. Once the action is complete, the system needs a pose estimate accurate enough to compare the model's position and orientation with a target pose.

On a standalone headset, camera processing, object tracking, and scene rendering share the same computing resources. This constraint motivates our preliminary investigation of low-compute, markerless tracking of the six-degree-of-freedom (6DoF) pose of a known physical model using the Meta Quest~3's rendered passthrough view for one eye and onboard processing.

\subsection{Design Goal and Contributions}

Our design goal is to estimate the pose of a known physical model from the passthrough view for one eye while leaving sufficient computing resources for mixed reality rendering on a standalone headset.

\begin{enumerate}
    \item \textbf{Synthetic training data.} We developed a Unity~\cite{unityEditor600034f1} pipeline that renders the teaching model under varied viewpoints, lighting, part configurations, occlusion, and motion blur. It automatically annotates 20 object keypoints and their visibility in each image.

    \item \textbf{On-device tracking.} We implemented a tracking pipeline on the Meta Quest~3 that updates a crop around the object and resizes it to a \(128 \times 128\) RGB input. A compact YOLO model trained on the synthetic dataset predicts the 2D keypoint locations. A perspective-\(n\)-point (PnP) solver estimates the object's pose for mixed reality visualization using these predictions and the corresponding known 3D points.

    \item \textbf{Preliminary evaluation.} We report 2D detection and
    keypoint localization results on synthetic validation images and
    runtime measurements on the Meta Quest~3. A qualitative headset
    demonstration shows how the estimated 6DoF pose drives a mixed
    reality overlay as a user rotates the physical airplane.
\end{enumerate}

\section{Related Work}

In BB8, Rad and Lepetit predicted the 2D projections of an object's
3D bounding-box corners and recovered its pose using
PnP~\cite{rad2017bb8}. Tekin et al. used a YOLO-inspired network to
predict these projections in a single forward pass~\cite{tekin2018singleshot}. We use the same geometric formulation with keypoints defined on the teaching model.

Tremblay et al. trained Deep Object Pose Estimation (DOPE) on a combination of photorealistic renders and domain-randomized images and used it to estimate the poses of known objects in real images~\cite{tremblay2018dope}. We use Unity to generate synthetic training images and keypoint annotations for a teaching model. The rendered scenes simulate conditions that arise as students handle the model, including changes in part configuration, hand occlusion, and motion blur.

Miu and Fryazinov demonstrated monocular 6DoF tracking on smartphones using one network to locate the object in the full image and a second network to track it within a moving crop~\cite{miu2020mobilepose}. Vuforia also demonstrates tracking of known objects by integrating Vuforia Model Targets with the device's camera feed~\cite{oquelisquforia}. Our system uses one keypoint model for both initial localization and subsequent tracking, expanding the crop when tracking is lost. The resulting pose estimates support visual feedback in AR-Classroom.

\section{System Design}
\subsection{Task Definition and Overview}

We estimate the 6DoF pose of a known LEGO airplane model from the Meta Quest~3's rendered passthrough view for one eye. The pose consists of a rotation \(\mathbf{R}\) and a translation \(\mathbf{t}\) that map points from the airplane's reference frame to the camera frame. The system uses the camera pose associated with each image to express the estimated object pose in the headset's tracking space. AR-Classroom can then use this pose to align virtual overlays with the airplane and compare its final pose with a target.

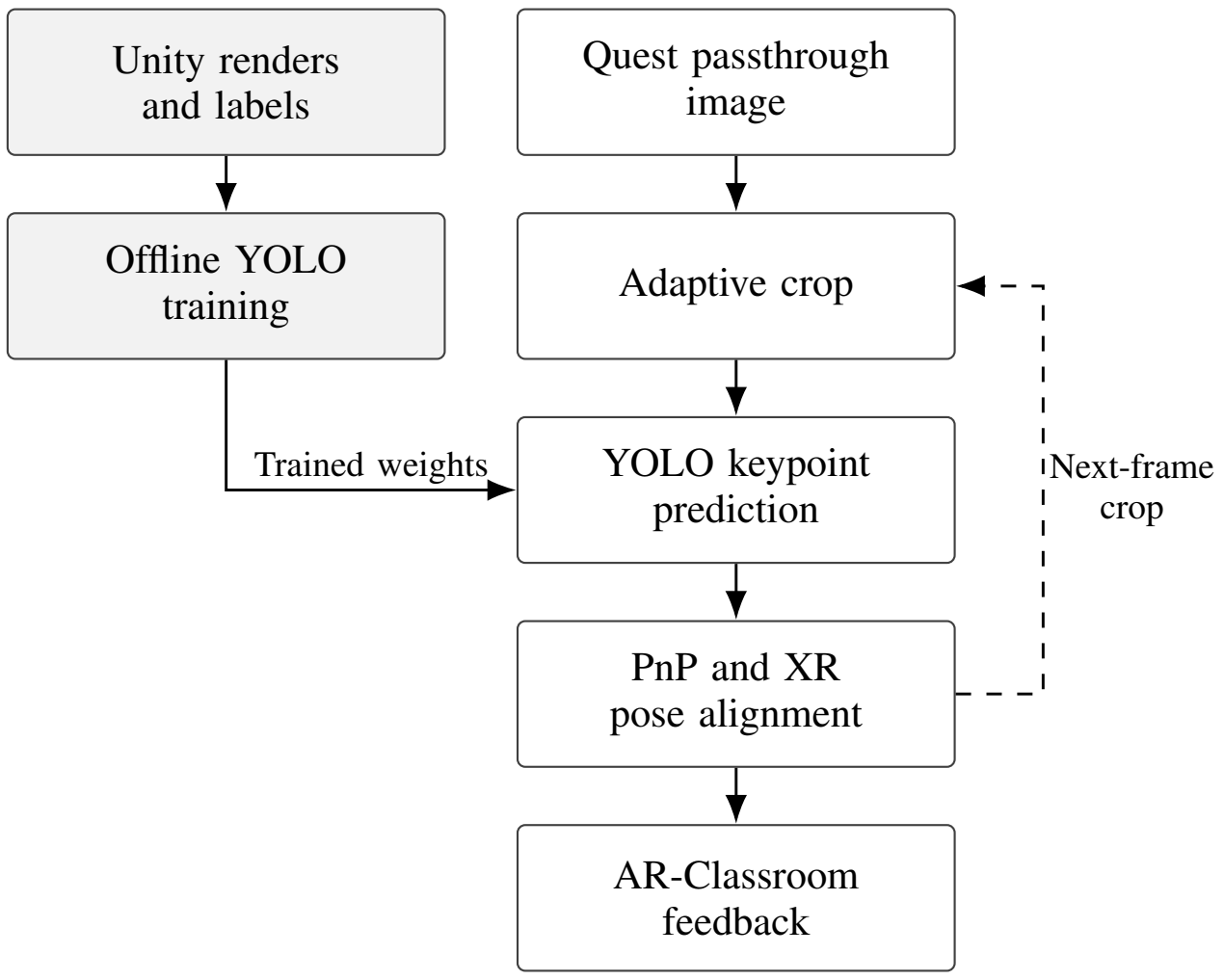
\begin{figure}[t]
    \centering
    \begin{tikzpicture}[
        x=1mm,
        y=1mm,
        font=\footnotesize,
        block/.style={
            draw=black!75,
            rounded corners=1.5pt,
            text width=26mm,
            minimum height=10mm,
            align=center,
            inner xsep=2mm,
            inner ysep=1.5mm
        },
        flow/.style={-{Latex[length=2mm]}, line width=0.55pt},
        flowlabel/.style={font=\scriptsize, align=center, inner sep=1pt}
    ]

    \node[block, fill=black!5] (unity) at (-35,0)
        {Unity renders\\and labels};
    \node[block, fill=black!5] (training) at (-35,-14)
        {Offline YOLO\\training};

    \node[block] (camera) at (0,0)
        {Quest passthrough\\image};
    \node[block] (crop) at (0,-14)
        {Adaptive crop};
    \node[block] (keypoints) at (0,-28)
        {YOLO keypoint\\prediction};
    \node[block] (pose) at (0,-42)
        {PnP and XR\\pose alignment};
    \node[block] (application) at (0,-56)
        {AR-Classroom\\feedback};

    \draw[flow] (unity.south) -- (training.north);
    \draw[flow] (camera.south) -- (crop.north);
    \draw[flow] (crop.south) -- (keypoints.north);
    \draw[flow] (keypoints.south) -- (pose.north);
    \draw[flow] (pose.south) -- (application.north);

    \coordinate (weightsTurn) at (training.south |- keypoints.west);
    \draw[flow] (training.south) -- (weightsTurn)
        -- node[flowlabel, above] {Trained weights} (keypoints.west);

    \coordinate (loopBottom) at ([xshift=6mm]pose.east);
    \coordinate (loopTop) at (loopBottom |- crop.east);
    \draw[flow, dashed] (pose.east) -- (loopBottom)
        -- node[flowlabel, right] {Next-frame\\crop} (loopTop)
        -- (crop.east);

    \end{tikzpicture}
    \caption{Offline training and on-device tracking of the LEGO airplane.
    The dashed arrow denotes the crop update for the next frame.}
    \label{fig:system-overview}
\end{figure}

Figure~\ref{fig:system-overview} shows the connection between offline training and on-device tracking. We render the airplane in Unity and generate keypoint annotations to train the YOLO model offline. On the headset, the tracker crops the current camera image and resizes the crop to a \(128 \times 128\) RGB input. The trained model predicts the image locations of the airplane's landmarks. The system maps these predictions back to the camera image coordinates and estimates the pose with PnP using the corresponding 3D landmarks and camera intrinsics. The detected bounding box and the validity of the pose estimate guide the crop for the next frame.

\subsection{Synthetic Data Generation and Training}

We generated 50{,}000 source images at \(640 \times 640\) pixels in Unity using a digital model of the LEGO airplane. To reflect the indoor settings in which learners handle the model, we drew approximately 90\% of the backgrounds from the classroom, living room, and library categories of the indoorCVPR\_09 dataset~\cite{quattoni2009indoor}. The remaining backgrounds were photographs taken in the project's intended use settings.

Following the principle of domain randomization~\cite{tobin2017domainrandomization},
we varied the airplane's position and orientation, lighting, part
removal and rotation, and motion blur. We placed capsule-shaped objects between the camera and the airplane to approximate the partial occlusion caused by fingers during handling. We set a 10\% probability of rendering a scene without the airplane. When the airplane was included, we set a 30\% probability of sampling a view near one of its six principal directions. These probabilities governed scene generation, so the realized proportions could vary across batches.

For scenes containing the airplane, we projected 20 predefined 3D landmarks into the image and checked whether each point fell within the frame and whether it was occluded. The standard YOLO Pose annotations were stored in separate text files containing the class ID, normalized bounding box, and landmark coordinates and visibility flags. Images without the airplane and images with fewer than two landmarks within the frame were retained as negative samples with empty annotation files.

We split the source renders into training and validation sets at an 80:20 ratio before preparing the cropped views. We retained full-frame examples and used a preset sampling rate of 80\% to generate additional crops around the annotated bounding boxes. All versions of a source render remained in the same split, and bounding boxes and landmark coordinates were remapped to the crop. The two types of views expose the model to the range of object scales encountered during wide-view search and close-up tracking on the headset. We resized the resulting images to \(128 \times 128\) pixels for RGB training of the YOLO26n-pose model~\cite{jocher2026yolo26}.

\begin{figure*}[t]
    \centering
    \includegraphics[width=\textwidth]{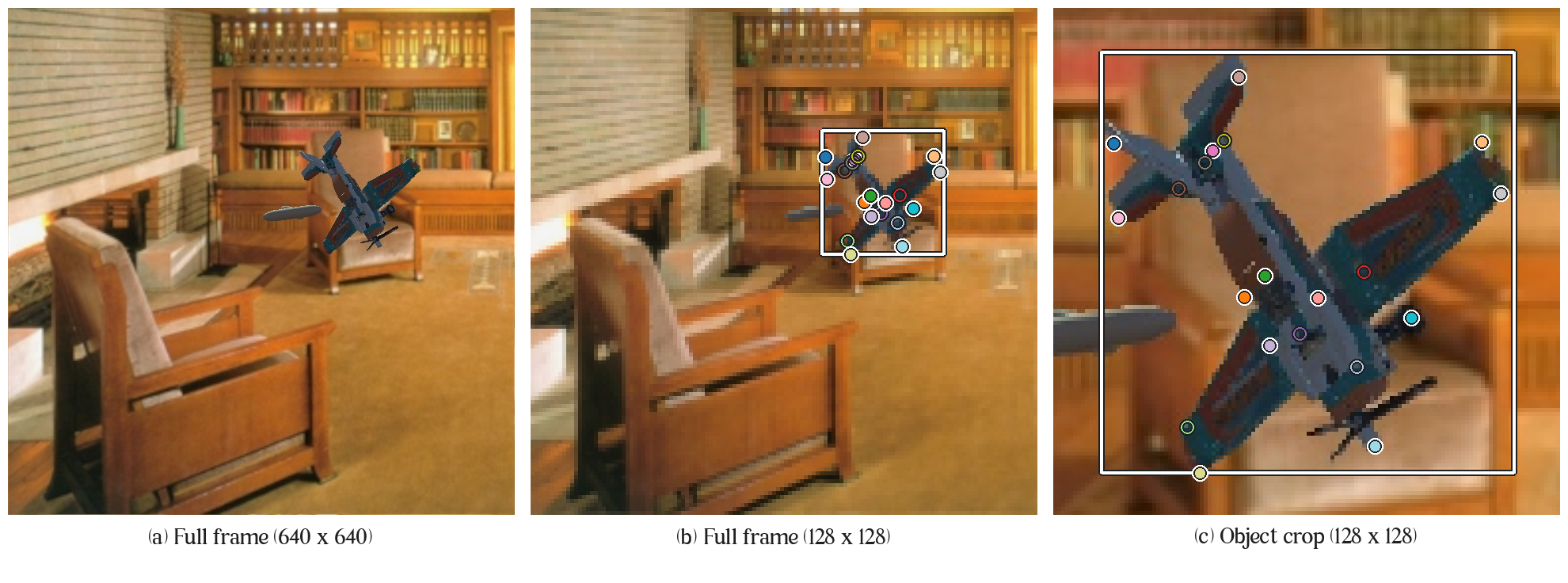}
    \caption{A synthetic render of the LEGO airplane and its two training views.
    Bounding boxes and keypoints are overlaid in (b) and (c) for visualization.
    Training annotations are stored in separate YOLO Pose label text files.}
    \label{fig:synthetic-training-views}
\end{figure*}

Figure~\ref{fig:synthetic-training-views} compares the original render with
the full-frame and cropped training views. At the same \(128 \times 128\)
input resolution, the airplane occupies more of the cropped view.
Additional synthetic examples are shown in the Appendix
(Fig.~\ref{fig:appendix-synthetic-examples}).

\subsection{Camera Input and Dynamic ROI}

The tracker starts with a wide search window in the RGB image from the Meta Quest~3's rendered passthrough view for one eye~\cite{metaPassthroughCamera}. When the tracker obtains a valid
pose, it uses the detected bounding box to set the region of interest
(ROI) for the next processed frame. The ROI follows the airplane as
tracking continues. If tracking is lost, the tracker progressively
expands the search window to reacquire the airplane.

The tracker resizes each crop to \(128 \times 128\) pixels for YOLO
inference on the headset using Unity Sentis
2.5.0~\cite{unitySentis250}. As illustrated by the synthetic example in
Fig.~\ref{fig:synthetic-training-views}(b) and (c), a tighter crop lets
the airplane occupy more pixels at the same input resolution. The
tracker records the crop origin and resize scale so that predicted
keypoints can be mapped back to the original camera image coordinates
for pose estimation.

\subsection{Pose Estimation and MR Feedback}

A target tracking rate of 2~Hz was the highest tested setting at which
the application maintained approximately 72~FPS, which is the typical required lowest application refresh rate for mixed reality applications.
Using the intrinsics for the same
image coordinates, we solve PnP with RANSAC~\cite{terzakis2020sqpnp,opencvPnP,enoxOpenCVForUnity} to estimate the airplane's
rotation and translation relative to the input view. We require at
least four inlier correspondences. We then use the view pose matched
to the image timestamp to transform the estimate into the XR tracking
frame.

The MR scene uses this pose to align a virtual model with the physical
airplane. To reduce visible jitter, we smooth the displayed position
and use quaternion interpolation for orientation. During a tracking
interruption, the virtual model holds its last displayed pose and is
hidden when the tracking-loss timer expires. AR-Classroom can use the
aligned pose to show how the airplane's local axes move as a learner
rotates it and to compare the final pose with a target pose.

\section{Preliminary Evaluation}

\subsection{Detection and Keypoint Localization on Synthetic Data}

After 150 training epochs, the standard validation run on 6,000
synthetic images yielded box mAP$_{50}$ and mAP$_{50\text{--}95}$
scores of 0.879 and 0.546, respectively. The corresponding 2D
keypoint scores were 0.678 and 0.440.

To examine the effect of cropping at a fixed input resolution, we
evaluated a 1,000-image subset of the synthetic validation split.
Of these images, 825 had airplane annotations and 175 had empty
labels. We processed each image both as a full frame and as a crop.
We used a crop centered on the ground-truth bounding box to crop for annotated images and a random
crop for images with empty labels. Both views were resized to
\(128 \times 128\) pixels and evaluated as RGB inputs.

We scored detection at the image level based on whether the model
returned a bounding box. A predicted keypoint counted as a hit if
its Euclidean distance from the corresponding annotation was at
most 6.4 input pixels. For each view, the hit rate was computed over landmarks labeled as
visible or occluded in that view. We computed the hit rate, precision,
and F1 for each landmark and averaged them across all 20 landmarks.

As shown in Table~\ref{tab:crop-comparison}, cropping increased the
median bounding-box long side from 57.1 to 106.8 pixels and raised
the detection rate on annotated images by 13.7 percentage points.
The mean keypoint hit rate also increased. Keypoint precision
decreased, and mean F1 improved by 1.7 percentage points. The
6.4-pixel tolerance was applied in model-input coordinates, so it
corresponded to a smaller distance in the source image for tighter
crops.

\begin{table}[t]
    \centering
    \caption{Paired comparison of full-frame and cropped
    \(128 \times 128\) inputs on 1,000 synthetic validation images.}
    \label{tab:crop-comparison}
    \small
    \begin{tabular}{lrr}
        \hline
        Metric & Full frame & Crop \\
        \hline
        Median GT box long side (px)       & 57.1  & 106.8 \\
        Detection rate on positives (\%)  & 32.5  & 46.2  \\
        Empty-label images with detections & 0/175 & 0/175 \\
        Mean keypoint hit rate (\%)       & 29.8  & 34.7  \\
        Mean keypoint precision (\%)      & 76.4  & 62.5  \\
        Mean keypoint F1 (\%)             & 42.9  & 44.6  \\
        \hline
    \end{tabular}
\end{table}

\subsection{On-Device Runtime and Interaction}
\label{sec:on-device-runtime}

We evaluated the \(128 \times 128\) RGB tracking pipeline on the
Meta Quest~3 at two target tracking rates. We recorded the application
frame rate (App FPS) and the number of completed tracking loops per
second (Loop FPS) to assess how increasing the tracking rate affected
the application's ability to maintain approximately 72~FPS.

At a target tracking rate of 2~Hz, the application stayed near 72~FPS.
When the target was increased to 6~Hz, the application ran at about
40~FPS and the tracking loop fluctuated between 5 and 6~Hz. The mean YOLO and PnP stage times were approximately 75–81 ms and 3.2–3.5 ms, and they were similar across the two settings.

\begin{figure}[t]
    \centering
    \begin{minipage}[t]{0.48\columnwidth}
        \centering
        \includegraphics[width=\linewidth]{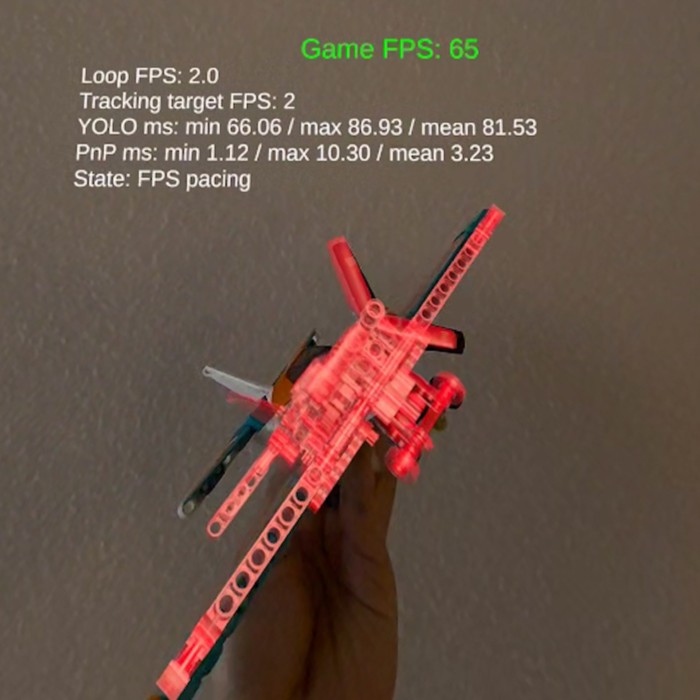}
        \par\smallskip
        {\footnotesize (a) Initial pose}
    \end{minipage}\hfill
    \begin{minipage}[t]{0.48\columnwidth}
        \centering
        \includegraphics[width=\linewidth]{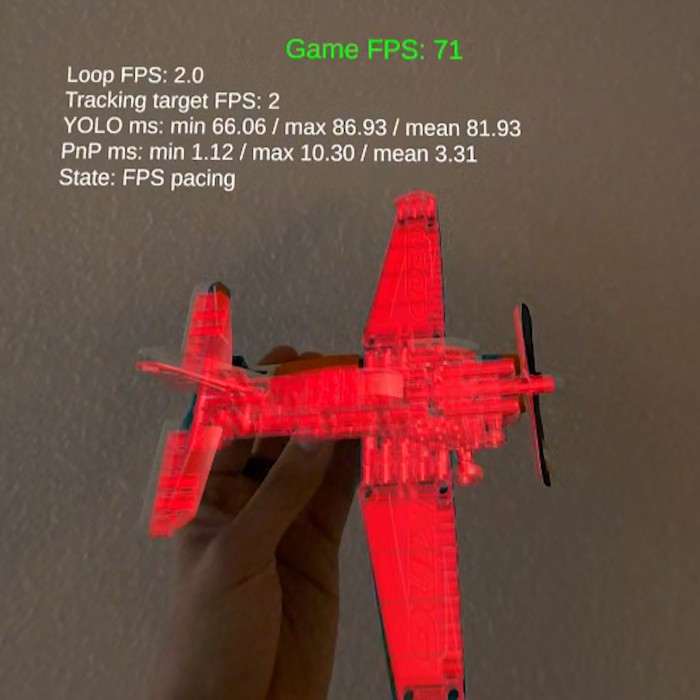}
        \par\smallskip
        {\footnotesize (b) Final pose}
    \end{minipage}
    \caption{Headset views from a single manual rotation on the
Meta Quest~3. The virtual overlay is shown after it has settled
at the airplane's initial pose (a) and final pose (b).}
    \label{fig:rotation-demo}
\end{figure}

Fig.~\ref{fig:rotation-demo} shows how the estimated 6DoF pose places
the virtual overlay at two orientations of the physical airplane.
The \href{https://youtube.com/shorts/ARABihD0x7E?feature=share}{%
    \textcolor{blue!65!black}{\underline{supplementary video}}}
shows the full rotation and the overlay's response during and after the motion.

\section{Discussion and Limitations}

At the same \(128 \times 128\) input resolution, cropping increased
the detection rate on annotated images by 13.7 percentage points
(Table~\ref{tab:crop-comparison}). The mean keypoint hit rate also
increased, alongside lower precision and a modest gain in mean F1.
On the Quest~3, 2~Hz was the highest tested tracking target at which
the application maintained approximately 72~FPS. At this setting,
the overlay receives new pose estimates infrequently during a rotation.

Across the two headset timing conditions, the YOLO stage averaged
approximately 75--81~ms, and PnP averaged 3.2--3.5~ms. This difference
makes YOLO the first target for reducing processing time. During
debugging, we often observed overlay misalignment alongside errors
in the predicted 2D keypoints. In the synthetic evaluation, the six
wing landmarks averaged about 38\% F1 in both views. The
other 14 landmarks averaged about 45\% in the full-frame view and
47\% in the cropped view. To improve wing and tail localization, we
plan to add an auxiliary training loss that supervises the separation
between left and right landmark groups and their positions relative
to fuselage reference points.

The quantitative localization results come from synthetic validation
images, with crops for annotated images derived from ground-truth
boxes. The headset demonstration illustrates 6DoF alignment
qualitatively for one physical object. We plan to use sequences with
reference poses to measure translation and rotation errors and
characterize recovery after tracking loss. To examine how 2D landmark
errors affect the estimated pose, we plan to run the same PnP solver
with annotated and predicted keypoints on the same frames and compare
both sets of estimates against the reference poses.

Moving from the desktop prototype to standalone XR involved reducing
model size and input resolution and using dynamic cropping to retain
more pixels on the airplane. Feeding the camera's native luminance
(Y) plane directly to a single-channel model could further reduce
preprocessing. We plan to compare this approach with RGB input under
the same headset conditions and measure end-to-end latency and
keypoint localization accuracy.

The two stable poses in Fig.~\ref{fig:rotation-demo} illustrate the
interaction supported by the current AR-Classroom prototype. A
learner can rotate the physical airplane and pause to inspect its
new orientation through the aligned virtual overlay.

\section{Conclusion}

We developed a markerless 6DoF tracker for a physical LEGO airplane
on the Meta Quest~3. The tracker uses a compact YOLO model trained
on synthetic Unity renders to predict 2D keypoints from dynamic
crops resized to \(128 \times 128\) pixels. It then estimates the
airplane's pose using PnP.

On synthetic validation images, cropping based on ground-truth boxes
increased the detection rate on annotated images by 13.7 percentage
points and raised mean keypoint F1 from 42.9\% to 44.6\%. On the headset,
2~Hz was the highest tested tracking target at which the application
maintained approximately 72~FPS. The rotation demonstration illustrates
how the prototype can support an AR-Classroom exercise in which a learner
rotates the physical airplane and pauses to inspect its new orientation
through an aligned virtual overlay.

\clearpage
\appendix

\begin{figure}[H]
    \centering
    \begin{minipage}[t]{0.48\columnwidth}
        \centering
        \includegraphics[width=\linewidth]{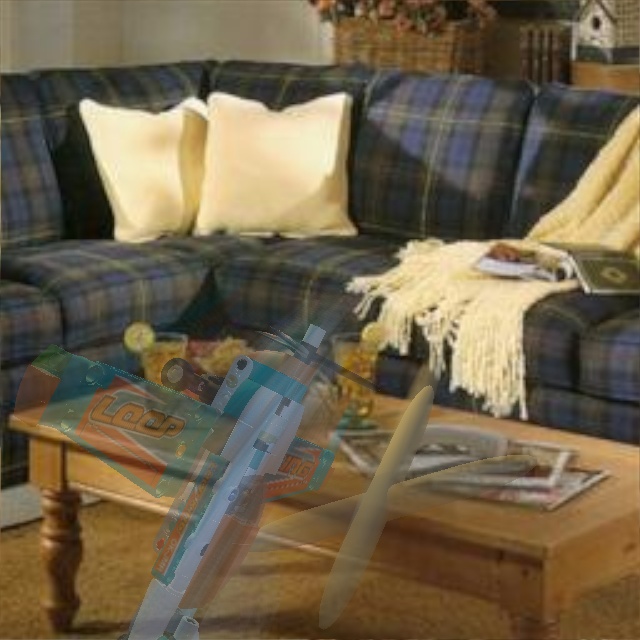}
        \par\smallskip
        {\footnotesize (a) Motion blur\par}
    \end{minipage}\hfill
    \begin{minipage}[t]{0.48\columnwidth}
        \centering
        \includegraphics[width=\linewidth]{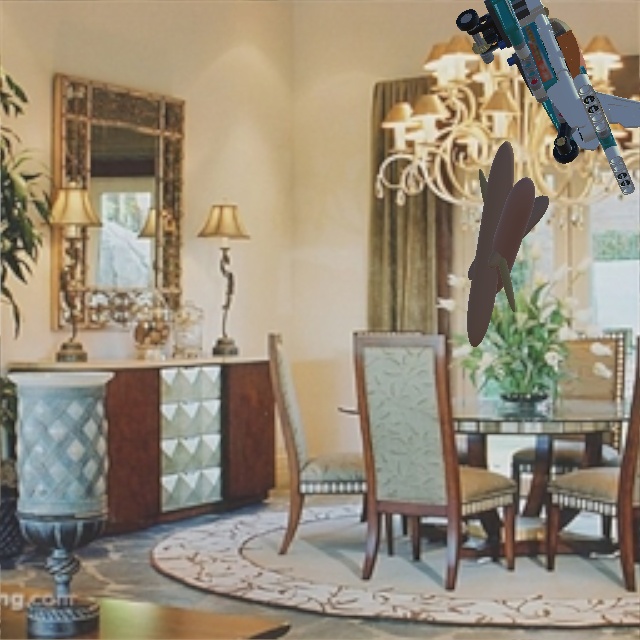}
        \par\smallskip
        {\footnotesize (b) Partially out of frame\par}
    \end{minipage}

    \par\medskip

    \begin{minipage}[t]{0.48\columnwidth}
        \centering
        \includegraphics[width=\linewidth]{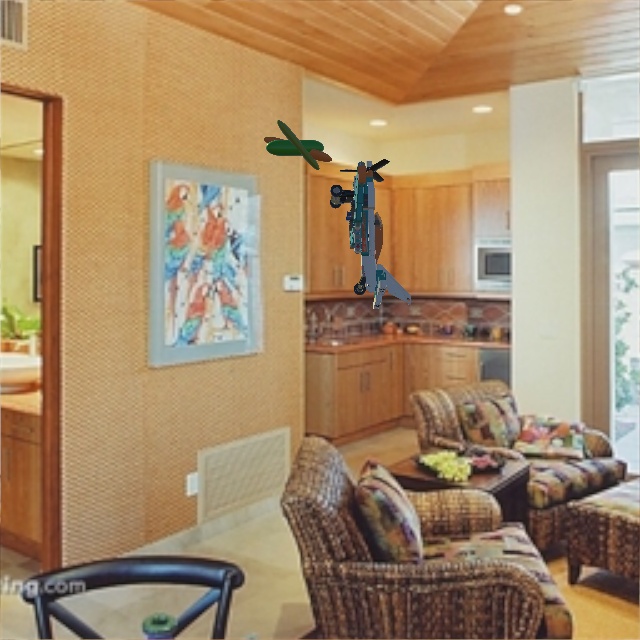}
        \par\smallskip
        {\footnotesize (c) Near-frontal view\par}
    \end{minipage}\hfill
    \begin{minipage}[t]{0.48\columnwidth}
        \centering
        \includegraphics[width=\linewidth]{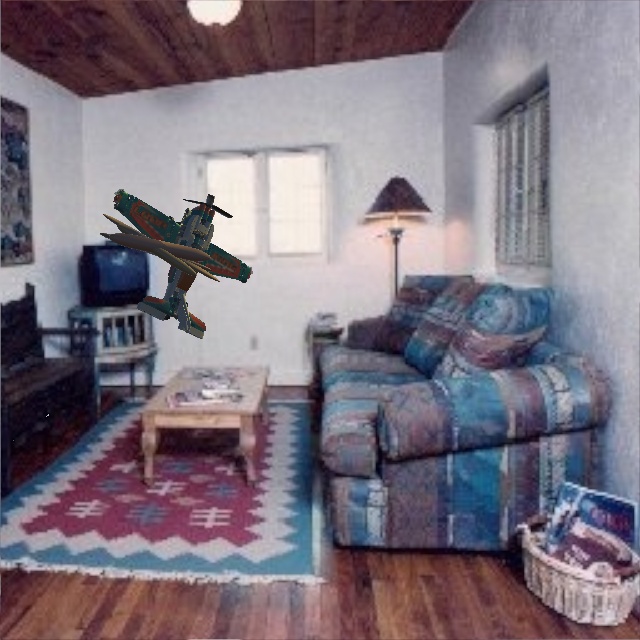}
        \par\smallskip
        {\footnotesize (d) Simulated finger occlusion\par}
    \end{minipage}

    \caption{Additional full-frame synthetic renders of the LEGO
    airplane generated in Unity.}
    \label{fig:appendix-synthetic-examples}
\end{figure}

\section*{Acknowledgment}
Portions of this research were conducted with the advanced computing resources provided by Texas A{\&}M High Performance Research Computing.

\bibliographystyle{IEEEtran}
\bibliography{references}

@article{shaghaghian2024spatial,
  author  = {Shaghaghian, Zohreh and Burte, Heather and Song, Dezhen and Yan, Wei},
  title   = {An augmented reality application and experiment for understanding and learning spatial transformation matrices},
  journal = {Virtual Reality},
  volume  = {28},
  year    = {2024},
  note    = {Art. no. 12},
  doi     = {10.1007/s10055-023-00904-x},
  url     = {https://link.springer.com/article/10.1007/s10055-023-00904-x}
}

@misc{arclassroom,
  author = {{AR-Classroom Project Team}},
  title  = {Project {AR-Classroom}},
  note   = {{Texas A\&M University}. Accessed: Nov. 12, 2026},
  url    = {https://sites.google.com/tamu.edu/ar-classroom/home}
}

@inproceedings{rad2017bb8,
  author    = {Rad, Mahdi and Lepetit, Vincent},
  title     = {{BB8}: A Scalable, Accurate, Robust to Partial Occlusion Method for Predicting the {3D} Poses of Challenging Objects without Using Depth},
  booktitle = {Proc. IEEE Int. Conf. Computer Vision (ICCV)},
  pages     = {3828--3836},
  year      = {2017},
  url       = {https://openaccess.thecvf.com/content_iccv_2017/html/Rad_BB8_A_Scalable_ICCV_2017_paper.html}
}

@inproceedings{tekin2018singleshot,
  author    = {Tekin, Bugra and Sinha, Sudipta N. and Fua, Pascal},
  title     = {Real-Time Seamless Single Shot {6D} Object Pose Prediction},
  booktitle = {Proc. IEEE Conf. Computer Vision and Pattern Recognition (CVPR)},
  pages     = {292--301},
  year      = {2018},
  url       = {https://openaccess.thecvf.com/content_cvpr_2018/html/Tekin_Real-Time_Seamless_Single_CVPR_2018_paper.html}
}

@inproceedings{tremblay2018dope,
  author    = {Tremblay, Jonathan and To, Thang and Sundaralingam, Balakumar and Xiang, Yu and Fox, Dieter and Birchfield, Stan},
  title     = {Deep Object Pose Estimation for Semantic Robotic Grasping of Household Objects},
  booktitle = {Proc. 2nd Conf. Robot Learning (CoRL)},
  series    = {Proceedings of Machine Learning Research},
  volume    = {87},
  pages     = {306--316},
  year      = {2018},
  url       = {https://proceedings.mlr.press/v87/tremblay18a.html}
}

@inproceedings{miu2020mobilepose,
  author    = {Miu, Valentin and Fryazinov, Oleg},
  title     = {Real-Time Monocular {6DoF} Object Pose Tracking on Smartphone {GPUs}},
  booktitle = {European Conf. Visual Media Production (CVMP)},
  year      = {2020},
  note      = {Short paper},
  url       = {https://www.cvmp-conference.org/files/2020/short/17.pdf}
}

@misc{oquelisquforia,
  author = {Oquelis, Daniel},
  title  = {{Quforia}: {Quest} + {Vuforia} Integration},
  note   = {{GitHub} repository. Accessed: Nov. 12, 2026},
  url    = {https://github.com/danieloquelis/Unity-QuestVuforia}
}

@inproceedings{quattoni2009indoor,
  author    = {Quattoni, Ariadna and Torralba, Antonio},
  title     = {Recognizing Indoor Scenes},
  booktitle = {Proc. IEEE Conf. Computer Vision and Pattern Recognition (CVPR)},
  pages     = {413--420},
  year      = {2009},
  doi       = {10.1109/CVPRW.2009.52065372},
  url       = {https://web.mit.edu/torralba/www/indoor.html}
}

@article{jocher2026yolo26,
  author  = {Jocher, Glenn and Qiu, Jing and Liu, Mengyu and Lyu, Shuai and Akyon, Fatih Cagatay and Kalfaoglu, Muhammet Esat},
  title   = {{Ultralytics YOLO26}: Unified Real-Time End-to-End Vision Models},
  journal = {arXiv preprint arXiv:2606.03748},
  year    = {2026},
  doi     = {10.48550/arXiv.2606.03748},
  url     = {https://arxiv.org/abs/2606.03748}
}

@inproceedings{tobin2017domainrandomization,
  author    = {Tobin, Josh and Fong, Rachel and Ray, Alex and Schneider, Jonas and Zaremba, Wojciech and Abbeel, Pieter},
  title     = {Domain Randomization for Transferring Deep Neural Networks from Simulation to the Real World},
  booktitle = {Proc. IEEE/RSJ Int. Conf. Intelligent Robots and Systems (IROS)},
  pages     = {23--30},
  year      = {2017},
  doi       = {10.1109/IROS.2017.8202133},
  url       = {https://arxiv.org/abs/1703.06907}
}

@misc{metaPassthroughCamera,
  author = {{Meta}},
  title  = {Passthrough Camera {API} Overview},
  note   = {{Meta Horizon OS Developers}. Accessed: Nov. 12, 2026},
  url    = {https://developers.meta.com/vr/documentation/unity/unity-pca-overview/}
}

@misc{unitySentis250,
  author = {{Unity Technologies}},
  title  = {{Sentis} Overview},
  note   = {{Unity Sentis documentation}, version 2.5.0. Accessed: Nov. 12, 2026},
  url    = {https://docs.unity3d.com/Packages/com.unity.ai.inference@2.5/manual/index.html}
}

@inproceedings{terzakis2020sqpnp,
  author    = {Terzakis, George and Lourakis, Manolis},
  title     = {A Consistently Fast and Globally Optimal Solution to the {Perspective-n-Point} Problem},
  booktitle = {Computer Vision -- ECCV 2020},
  series    = {Lecture Notes in Computer Science},
  volume    = {12346},
  pages     = {478--494},
  year      = {2020},
  doi       = {10.1007/978-3-030-58452-8_28},
  url       = {https://www.ecva.net/papers/eccv_2020/papers_ECCV/papers/123460460.pdf}
}

@misc{opencvPnP,
  author = {{OpenCV}},
  title  = {{Perspective-n-Point} ({PnP}) Pose Computation},
  note   = {OpenCV 4.x documentation. Accessed: Nov. 12, 2026},
  url    = {https://docs.opencv.org/4.x/d5/d1f/calib3d_solvePnP.html}
}

@misc{enoxOpenCVForUnity,
  author = {{Enox Software}},
  title  = {{OpenCV for Unity} Documentation},
  note   = {Accessed: Nov. 12, 2026},
  url    = {https://enoxsoftware.com/opencvforunity/documentation/}
}

@misc{unityEditor600034f1,
  author = {{Unity Technologies}},
  title  = {{Unity Editor 6000.3.4f1} Release Notes},
  year   = {2026},
  note   = {Accessed: Nov. 12, 2026},
  url    = {https://unity.com/releases/editor/whats-new/6000.3.4f1}
}

\end{document}